\documentclass[conference]{IEEEtran}
\usepackage{amsmath}
\usepackage{hhline}
\usepackage[dvips]{graphicx} 
\usepackage{bmpsize} 
\usepackage{epsfig}
\usepackage{multirow}
\usepackage{setspace}
\usepackage{color}
\usepackage{url}

\usepackage{enumerate}
\usepackage{amsfonts}
\usepackage{amsthm}
\usepackage{amscd}
\usepackage{amssymb}

\newtheorem{definition}{Definition}

\begin{document}

\title{Watchlist Risk Assessment using Multiparametric Cost and Relative Entropy}

\author{\IEEEauthorblockN{K. Lai and S.N. Yanushkevich}
\IEEEauthorblockA{Biometric Technologies Laboratory, Department of Electrical and Computer Engineering\\
University of Calgary, Alberta, T2N 1N4 Canada\\ 
Email: \{kelai,syanshk\}@ucalgary.ca}}


\markboth{IEEE SSCI,~2017}{ \MakeLowercase{\textit{et al.}}:
.....}

 \maketitle
\IEEEoverridecommandlockouts
\IEEEpubid{\begin{minipage}{\textwidth}\ \\[55pt]
		\footnotesize{{\fontfamily{ptm}\selectfont Digital Object Identifier 10.1109/SSCI.2017.8285219 \\978-1-5386-2726-6/17/\$31.00 \copyright 2017 IEEE}}
\end{minipage}}

\begin{abstract}
This paper addresses the facial biometric-enabled watchlist technology in which risk detectors are mandatory mechanisms for early detection of threats, as well as for avoiding offense to innocent travelers.  We propose a multiparametric cost assessment and relative entropy measures as risk detectors.  We experimentally demonstrate the effects of mis-identification and impersonation under various watchlist screening scenarios and constraints. The key contributions of this paper are the novel techniques for design and analysis of the biometric-enabled watchlist and the supporting infrastructure, as well as measuring the impersonation impact on e-border performance.
\end{abstract}

\vspace{0.4cm}
\begin{IEEEkeywords} \textit{Biometrics, automated border control, watchlist, risk assessment, entropy.   }
\end{IEEEkeywords}

\section{Introduction}
One of the urgent problems is applying computational intelligence techniques to address the scenario of travelers passing through the automated borders \cite{[EC-SmartBorders-2014]}. In this scenario, the two primary functions of the Automated Border Control (ABC) machine are to perform the following (under the time constraint of several minutes):  1) authentication by using the traveler's appearance and e-passport (most often the e-passport will include a facial template), and 2) assessment of risk through the use of a watchlist. A contemporary watchlist screening technology is unreliable and inefficient \cite{[US-Department-Justice-Office]}. This is because the watchlist contains only alphanumeric data (name, data of birth, etc.) which can be altered or forged. One possible solution for this problem is to add biometric traits of the person of interest into the watchlist. The biometric technologies have been proven to be acceptable in entry-exit concepts for visa visitors \cite{[EC-SmartBorders-2014],[DHS(2010)]}, but were less successful for ABC machines due to unacceptable performance degradation when automation is applied to decision making. Contemporary ABC machines direct on average 3 out 10 people to manual control \cite{[Eastwood-IEEE-J-2015]}.  If a facial-enabled watchlist is integrated into the system, 5 or more out of 10 people may be directed to the border officer which leads to significant degradation of the system's performance.

There are various approaches to solve this  problem. One approach is to include the use of the biometric menagerie or Doddington phenomenon \cite{[kn:Doddington-1998]}.  The Doddington phenomenon studies the effects of impersonation \cite{[Bustard-Nixon-2013],[Yager-2010]} and suggest the use of multi-biometrics \cite{[Poh-Kittler-Bourlai-2010]} to alleviate the problem. However, in the current ABC machines, the people of interest are mostly represented by facial traits from the physical and digital world, and are less often  to be identified by fingerprints or irises \cite{[Best-Rowden-2014]}. Another approach is to improve the quality of facial traits used in the watchlist \cite{[Bourlai-2011]}. The improvement of the image quality can be used to mitigate the effects of plastic surgery \cite{[Jillela-2012]}, makeup \cite{[Chen-2013]}, spoofing \cite{[Maatta-2012]}, and age progression \cite{[Geng-age-estimation-2007],[Hunter-facial-ageing-2012]}. Ideally, the watchlist should contain  synthetic facial images  of people of interest; which can be created by composite machines.  Finally, the development of better algorithms, such as deep learning \cite{[Szegedy-2015]}, shall increase the performance and accuracy of the facial recognition algorithm used by the watchlist system.  

In practice, the main obstacle in achieving an acceptable solution is the nature of the watchlist: the watchlist contains mostly low quality facial traits which are the sources of intentional or un-intentional impersonation, and thus, false acceptance or rejections. There are various techniques to mitigate these sources such as mandatory quality management \cite{[Spreeuwers-Face-Schiphol-Airport]},  but this condition is not always possible to create.       

The aforementioned problems explain the reason why biometric-enabled watchlist is still not integrated into the border control systems. There is an urgency in finding a solution to this problem, as the use of biometrics may cause more mismatch cases compared with today's  non-biometric practice of watchlist check in which the key operation to address mismatches is the Redress Complaint Disposition  \cite{[US-Department-Justice-Office]}. 
However, any  watchlist check addresses the risk that an innocent person is identified as a person of interest, and vice versa, a person of interest is missed. Conceptually, the solution exists in the form of inference engine over the landscape of particular solutions such as Doddington menagerie. From this perspective, the watchlist screening must include the assessment of impersonation risk before making any decisions.

The above is the motivation of using biometric-enabled watchlist technology to construct a system that works under time constraints and specific reliability requirements. 

In  Section 2, we provide the key definitions and details
of the problem. Our results are introduced in Section 3.
In Section 4, we discuss the contribution of this paper and
formulate conclusions.


\section{Basic definitions and problem statement}\label{sec:Background}

The  watchlist technology includes the following key components:  update mechanisms;  architecture (usually a distributed network);
 performance metrics (including social impact); social embedding (capacity of sources of information);  watchlist inference (engine, capacity, and control); and data protection mechanism.


For the watchlist technology, there are two kinds of risk: 

1)  Risk of mis-identification caused by the watchlist, and
	
2)  Risk of mis-identification in the course of the watchlist check for a given traveler.

  These different kinds of risk can be addressed in terms of levels.

\begin{definition}
 {\bf Risk level}  is defined by the following scenarios in the watchlist application:\\
Level-I (host system): Given a watchlist, the potential risk for an arbitrary traveler is defined by the distribution of probability of error over the data structure in an appropriate metric. \\
Level-II (authentication and risk assessment station): Given a traveler, the risk of his/her mismatch against a  watchlist is defined with respect to the potential risks of this watchlist.
 \end{definition}

In the host system, Level-I risks are periodically updated and evaluated to create a ``risk landscape''.  The `risk landscape'' is then presented to the authentication and risk assessment station located at the border checkpoint, e.g. in airport. At Level-II, the watchlist risk landscape is applied to a given traveler once his/her biometric traits are acquired.    


Risk is generally understood as a random event that has unwanted consequences when the event occurs.
 Risk events are analyzed in terms of their probability of occurrence and their impact on the system performance:  \cite{[kn:Rausand-2011]}:
$${{\texttt{Risk} = F(\texttt{Probability, Impact})}}$$
For risk description, the form called \emph{risk statement} is used. \emph{Risk score} can be defined as a numerical value,  or can be presented in a semantic form such as ``Almost/extremely sure to occur'', ``Likely/unlikely to occur'', and ``Somewhat less/greater than an event chance to occur''.

There are many sources of risk for the biometric-enabled watchlist.  One type revolves around face identification across age progression, facial hair style, accessories, facial implants and surgery. Another group of risks is related to the architectural and implementation aspect of the watchlist system, such as compromised server data and malicious actions of server personnel.


Watchlist technology can be evaluated using a wide variety of metrics, including risks, costs, and benefits. For example, direct benefits include reduced costs for traveler risk assessment, and prevented/mitigated attacks during execution.
Indirect benefits include lesser chances for successful attacks and improved public perceptions of e-borders security.
 
One of the metrics, that we suggest to apply to the aforementioned scenarios is the relative entropy, or Kullback-Leibler (KL) measures, which are suggested to be useful in pattern analysis for analyzing information.
In \cite{[Zouaouia-2015]}, KL measures are used in the shout detectors in a railway carriages for improving the security for passengers. The shout detector aims at estimating the probability that a sound signal will contain some segments of a shout. The KL divergence is used as the measure of similarity between the class ``shout'' signals and class ``background'' noise.  In addition, we refer to papers \cite{[Lim-2016],[Liu-2011],[Zhang-2011]} where other information-theoretical measures have been used for resolving  problems related to image analysis and recognition.

This paper addresses the risks of application of watchlists containing facial images of persons of interest. Compared to traditional watchlists used in ABC machines,
biometric watchlist are characterized by additional risks. These risks should be detected before making any decision. Such detectors are intended for early manifestation of potential threats as well as for reducing or avoiding the possibility of treating an innocent traveler as a suspect. In addition, these detectors should operate with various data structures because of various nature of risks.

The key source of impersonation is biometric diversity which is a robust empirical phenomenon that makes a perfect separation into classes in the recognition process  hardly possible.  Sources of impersonation include the following  \cite{[Eastwood-IEEE-J-2015],[Poh-Kittler-Bourlai-2010]}: 
(a) permanent and  temporary biometric abnormalities, (b) imperfection of the biometric acquisition techniques, and
(c) algorithmic recognition differences and imperfections. The main countermeasures for impersonation, or \emph{conditions of immunity}  to impersonation, is to improve the quality of both the biometric samples and refine the pattern recognition algorithm. 

In e-borders, impersonation is specified by the various factors such as recognition mode (verification for e-passport holders or/and identification for risk assessment via watchlists), carrier  (semantic in interview supported machines or biometric data in ABC machines), depth of embedding in social infrastructure, and privacy issue (how deeply national legislation allows to profile travelers for risk assessment).

	%
	
	An appropriate metric for the biometric-enabled recognition  known as \emph{Doddington categorization} is well studied for various modalities \cite{[kn:Doddington-1998],[Poh-Kittler-Bourlai-2010]}, including the database which we used in experiments \cite{[Witt2006]}, as well as the extended Doddington categorization \cite{[Yager-2010]}.
	
\begin{definition}\label{def:Doddington}
{\bf Doddington  metric} is defined as the four type classification of recognition process:  
 Category I (`sheep'), recognized normally (have high genuine scores and low impostor scores);	
 Category II (`goats'),  hard to recognize (have low genuine scores);
 Category III (`wolves'), good at impersonating (have high impostor scores); and
 Category IV (`lambs'),  easy to impersonate (have high genuine and impostor scores). 
\end{definition}
The process of determining the Doddington categories is based on analyzing the order of the recognition scores ('goats' for genuine and 'wolves' and 'lambs' for imposter).  In this paper, 2.5\% of the population is selected as the 'goats', `wolves' and `lambs' categories because 2.5\% signifies two standard deviations from the mean.  In the case of our experimental setup, `goats' are 15 individuals (out of 568) that obtained the lowest genuine scores. Similarly, the  'wolves' and 'lambs' are 15 individuals (out of 568) that achieved the highest imposter scores. Finally, 'sheep' are individuals that neither belong to the 'goat' nor the 'wolf' categories.

The extended Doddington's categorization \cite{[Yager-2010]} include 'doves', 'chameleons', 'ghosts', and 'worms'.  Later in the paper we will use 'worms', which are individuals exhibiting the features of both 'goats' (have low genuine scores) and 'wolves' (high impostor scores).

\begin{figure}[!hbt]
\begin{center}
		\begin{small}
\begin{tabular}{cccc} 
\includegraphics[width=0.11\textwidth]{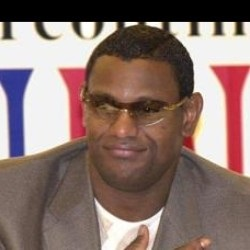}&
\hspace{-3mm}\includegraphics[width=0.11\textwidth]{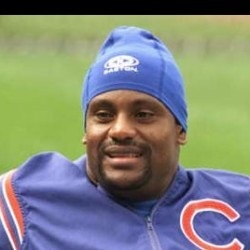}&
\hspace{-3mm}\includegraphics[width=0.11\textwidth]{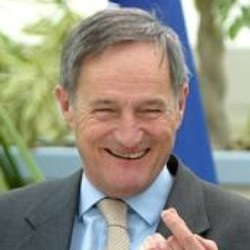}
&
\hspace{-3mm}\includegraphics[width=0.11\textwidth]{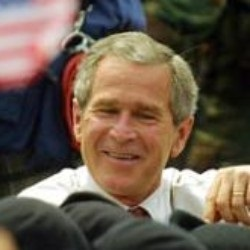}
\\
$(a$) & $(b)$ & $(c)$ & $(d)$
\end{tabular}
\end{small}
\end{center}
 \caption{Images of subjects in Doddington metric:
  $(a)$ and $(b)$ are  'goat' subject;  $(c)$ an image of a 'wolf/lamb' subject that looks similar to subject $(d)$  (images are from the LFW database \cite{LFWTech}).
}\label{fig:photosLFW}
  \end{figure}

Typical examples of the risk of the watchlist screening, using Doddington  metric are   given in Fig. \ref{fig:photosLFW}:  left pair of images results in mis-identification, and the right pair illustrates the impersonation effect.

Doddington's phenomenon is not stable and depends on many factors \cite{[Yager-2010]}. Doddington's detector utilizes  two classes of match scores: \emph{genuine}  and \emph{impostor} match scores.
\begin{definition}
{\bf Genuine} and {\bf impostor} match scores are derived from  comparing two biometric samples belonging to the same and different individuals, respectively.
\end{definition}

Our experiments concern with the following scenarios of the border crossing passage: 
\begin{itemize}
	\item [] \hspace{-7mm}\textbf{Scenario 1:} The traveler is a person of interest, he/she belongs to the class of 'goats' and, thus, poses a risk of not being matched against the watchlist, and, consequently, passing through the border control without being detected.	
	\item  []\hspace{-7mm}\textbf{Scenario 2:} The traveler is not a person of interest,  he/she belongs to the class of  `lambs' (and `wolves' in a symmetric matcher) and thus, might match against someone on the watchlist and generate a false alert and the likely RCD (Redress Complaint Disposition) procedure; 
	\item []\hspace{-7mm}\textbf{Scenario 3:} The traveler is a person of interest, he/she belong to the class of  `wolves' and may create a lot of matches against the watchlist, thus, causing logistic problems; however, this case may not be what the attacker wants as this event will still alert border control. 
\end{itemize}  

In the current ABC machines, different kinds of biometrics are used for identification including fingerprints, faces, iris, and/or retina.  Each biometric has been shown to behave independently of each other, thus each biometric will provide their own unique Doddington categories. A study was performed in \cite{Lai-Watchlist-2017} that shows the risk calculated using face and fingerprint biometrics are unique.

In this paper, we chose to use facial databases to simulate the watchlists because faces are one of the most common and least invasive biometrics. Databases FRGC V2.0 and Faces-in-the-Wild (LFW) \cite{LFWTech} contain facial images of various quality. We used commercially available recognition software Verilook for facial images  (one commercial software examined in the FRVT 2013 NIST competition \cite{frvt2013}), from Neurotechology.

\section{Watchlist risk landscape}\label{sec:}

In this Section, we introduce two types of risk detectors. The first one is using the  idea of tracking a set of parameters (error rates) which are related to various sources of risks. We called this process a multiparametric cost assessment. The idea of second detector is inspired by information theory.

\subsection{Detector I: Multiparametric cost assessment}

Our goal is to estimate risks for decision-making using Doddington's categorization and multiparametric cost functions. We define the cost function using the following control parameters of unstable Doddington phenomenon: (a) the threshold and (b) the cost of false rejection and false acceptance errors.

\begin{definition}
Watchlist screening or negative identification, establishes whether a traveler is not on the watchlist. It is characterized by {\bf False Negative (FN)}  (miss-match, or miss-identification) errors and {\bf False Positive (FP)}, or False Alarm   (false detection, or impersonation).
\end{definition}

The corresponding error rates are called a False Negative Rate (FNR) and  a False Positive Rate (FPR) \cite{[kn:Bolle04]}. 
 A FP results is a convenience problem, since an innocent traveler is denied access and needs to be manually checked or examined to get access. 

It is documented that an expected overall error rate, $E(T)$, of a general recognition system can be estimated using likelihood of both rates given a threshold, $T$,  as well as the prior probability $P_I$  of  a random user being an impostor and the prior probability $P_G$ of a user being genuine \cite{[kn:Bolle04]}: 
\begin{equation}\label{eq:overall-error-rate}
E(T) =  \texttt{FAR}(T) \times P_I + \texttt{FRR}(T) \times P_G
\end{equation}
where the $FRR$ is False Rejections Rate  and $FAR$ is False Acceptance Rate.
Equation \ref{eq:overall-error-rate} represents a probability that a random trial will result in an error for that particular matcher given threshold $T$. 

The overall expected error (Equation \ref{eq:overall-error-rate}) can be used to define a cost associated with each error, such as the cost of false positives, $C_{FP}$, and the cost of false negatives, $C_{FN}$. 
For a watchlist, the expected cost of each match decision  is evaluated by Equation \cite{[kn:Bolle04]}:
\begin{equation}\label{eq:cost}
\texttt{Cost} = C_{FN} \times \texttt{FNR} \times P_G + C_{FP} \times \texttt{FPR} \times (1-P_G)
\end{equation}

Consider a watchlist that contains biometric data on $n$ people;  let $m$ be the number of travelers passing through the border control point (denoted by $\texttt{FPR}(1)$).  Let the cost of a wanted person to be mis-identified and the cost of an unnecessary `bottleneck' be $C_{\texttt{FN}}$ and $C_{\texttt{FP}}$, respectively.

\begin{definition}
Risk of the watchlist screening is defined as the overall error rate with respect to a certain threshold $T$ \emph{\cite{[kn:Bolle04]}}:
\begin{equation}
\emph{\texttt{Risk}}(T) = \emph{\texttt{FNR(T)}} + \frac{{ C_{{\texttt{FN}}}} \times (1- P_G)} { C_{{\texttt{FP}}} \times P_G }\times \emph{\texttt{FPR}} (T) 
\label{eq-risk}
\end{equation}
\end{definition}

 In Doddington's metric, the category of `goats' is associated with high rejection rate. We will attribute high $FNR$ to the 'goats' and 'worms' categories. Also, the category of `wolves/lambs', as well as 'worms', contributes to $FPR$.
Let us assume that the cost of mis-identification of a wanted person is 10 times higher than the cost of mismatch of a regular traveler, 
$C_{\texttt{FN}}= 10 \times C_{\texttt{FP}}$; given $P_G=0.1$.

$$\texttt{Risk}(T) = \texttt{FNR} (T) + 90 \times \texttt{FPR}(T)$$

Let us denote Doddington category as $D_i$, where $i$ indicates the  category such that $D_1$ is 'sheep', $D_2$ is 'goat', $D_3$ is 'wolf/lamb' and $D_4$ is 'worm',
 Prior probability of a random traveler being genuine, $P(G,D_i)$, and imposter, $P(I,D_i)$, depends on the prior probabilities of that traveler belonging to $i$-th Doddington category, $D_i$:
\begin{eqnarray} 
	P(G,D_i)&=&P(G)\times P(D_i)\label{eqn.pgi-1}\\
	P(I,D_i)&=&P(I)\times P(D_i)\label{eqn.pgi-2}
\end{eqnarray}

Finally, using Equations \ref{eqn.pgi-1} and \ref{eqn.pgi-2}  the Doddington landscape, obtained using the Doddington metric defined in Definition 2, can be defined in terms of cost  assessment:  
\begin{eqnarray} 
\texttt{Cost}_i(G)&=&C_{FN}	\times \texttt{FNR}_i \times P(G,D_i) \label{eq:cost-final-1}\\
\texttt{Cost}_i(I)&=&C_{FP}	\times \texttt{FPR}_i \times P(I,D_i) \label{eq:cost-final-2}
\end{eqnarray}

Equations \ref{eq:cost-final-1} and \ref{eq:cost-final-2} states that: 

1) The scenarios for genuine and imposter traveler should be distinguished;

2) The cost of these scenarios can be controlled by security personnel via weights $C_{FN}$ and $C_{PN}$;

3) The $FNR$ and $FPR$ impact the cost directly;

4) The prior probabilities ($P(G,D_i)$ and $P(I,D_i)$) control the amount of impact a random traveler put on the cost, given his genuine/imposter identity and assigned Doddington category.

5) The cost of assessment is the product of the cost of error ($C_{FN}$ and $C_{PN}$), the chance of error ($FNR$ and $FPR$), and the influence caused by the specific traveler ($P(G,D_i)$ and $P(I,D_i)$, respectively).

Table \ref{tab:risk2} contains the estimation of risk in terms of a cost function, FNR and FPR for three thresholds $T=10,50, 100$ for various Doddington classes. We can conclude that:

1) The threshold $T$ impacts the value of risk.  Increasing the value of the threshold will increase the $FNR$ risk and decrease the $FPR$ risk.
	
2) A `goat' can influence the cost of $FN$ while a `wolf' can impact the cost of $FP$.  A worm, exhibiting the features of  both a 'goat' and a 'wolf', can control the cost of both $FN$ and $FP$.
		
3) The quality of images in a database can influence the impact a chosen set of thresholds can cause. A database with high quality images (FRGC) will generate higher match scores than a database with low quality (LFW) images.

\begin{table}[!htb]
\caption{Watchlist risk landscape in terms of  cost metric with respect to the two risk classes for various thresholds. }	\centering
		\begin{footnotesize}
		\begin{tabular}{|l||l|l|l|l|}
		\hline
			 &  \multicolumn{1}{|c|}{"FNR Risk"}  &  \multicolumn{1}{|c|}{"FNR Risk"}  &  \multicolumn{1}{|c|}{"FPR Risk"}  &  \multicolumn{1}{|c|}{"FPR Risk"} \\
			&  \multicolumn{1}{|c|}{Goat}&   \multicolumn{1}{|c|}{Worm} &   \multicolumn{1}{|c|}{Wolf} &  \multicolumn{1}{|c|}{Worm}\\
	 &  \multicolumn{1}{|c|}{($\times C_{FN}$)}	 &   \multicolumn{1}{|c|}{($\times C_{FN}$)}	&   \multicolumn{1}{|c|}{($\times C_{FP}$)}	 & \multicolumn{1}{|c|}{($\times C_{FP}$)}	\\
		\hline
		\hline
		T&\multicolumn{4}{|c|}{FRGC database } \\
		\hline
	10	& 1.70$\times 10^{-7}$&	0	&	0	&	0	\\
		50	& 1.01$\times 10^{-5}$&	0	&	0	&	0	\\
		100	& 2.31$\times 10^{-5}$	&	0	&	0	&	0	\\
		\hline
		\hline
		T&\multicolumn{4}{|c|}{LFW database } \\
	\hline
		10	& 1.45$\times 10^{-5}$	&	7.22$\times 10^{-7}$ &	1.19$\times 10^{-3}$	&	2.40$\times 10^{-2}$	\\
		50	& 1.45$\times 10^{-5}$ &	7.22$\times 10^{-7}$	&	0	&	0	\\
		100	& 1.45$\times 10^{-5}$&	7.22$\times 10^{-7}$ 	&	0	&	0	\\
	\hline
		
					\end{tabular}
					\end{footnotesize}
	\label{tab:risk2}
\end{table}

\subsection{Detector II: Relative entropy metric}
It can be shown that Kullback-Leibler (KL) divergence, or relative entropy can be useful measure for evaluating the Doddington's landscape given the similarity of probabilistic distributions. The relative entropy metric is defined by the equation \cite{[Kullback-Leibler-1951]}: 
\begin{eqnarray*}
\text{Relative entropy} =D(P||Q)=\sum_{j}p_{j}\log_2\frac{p_j}{q_j}
\end{eqnarray*}
where $P=[p_1,p_2,\ldots p_n]$ and $Q=[q_1,q_2,\ldots q_n]$ are probabilistic distributions.  The KL divergence is a positive value,  $D(P||Q)\geq 0$. If the distributions are the same, then the KL divergence is zero; the closer they are, the smaller the value of $D(P||Q)$. Hence, the divergence can be computed for any pair of Doddington's categories.

\begin{figure}[!hbt]
\begin{center}
\begin{small}
\begin{tabular}{cc} 
\begin{parbox}[h]{0.5\linewidth}{\centering 
\includegraphics[width=0.265\textwidth]{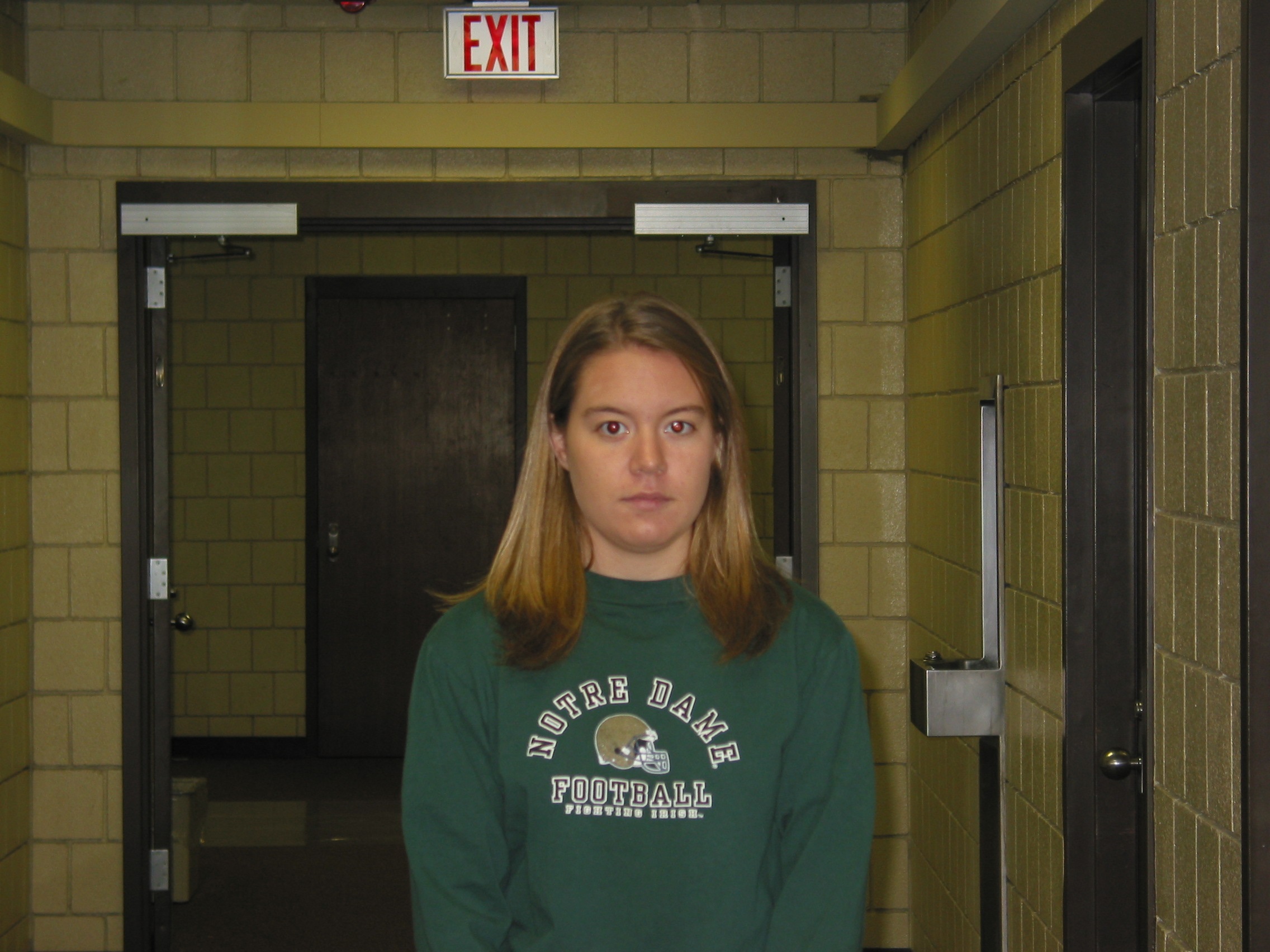}
}\end{parbox}
&
 \begin{parbox}[h]{0.4\linewidth}{\centering 
\includegraphics[width=0.15\textwidth]{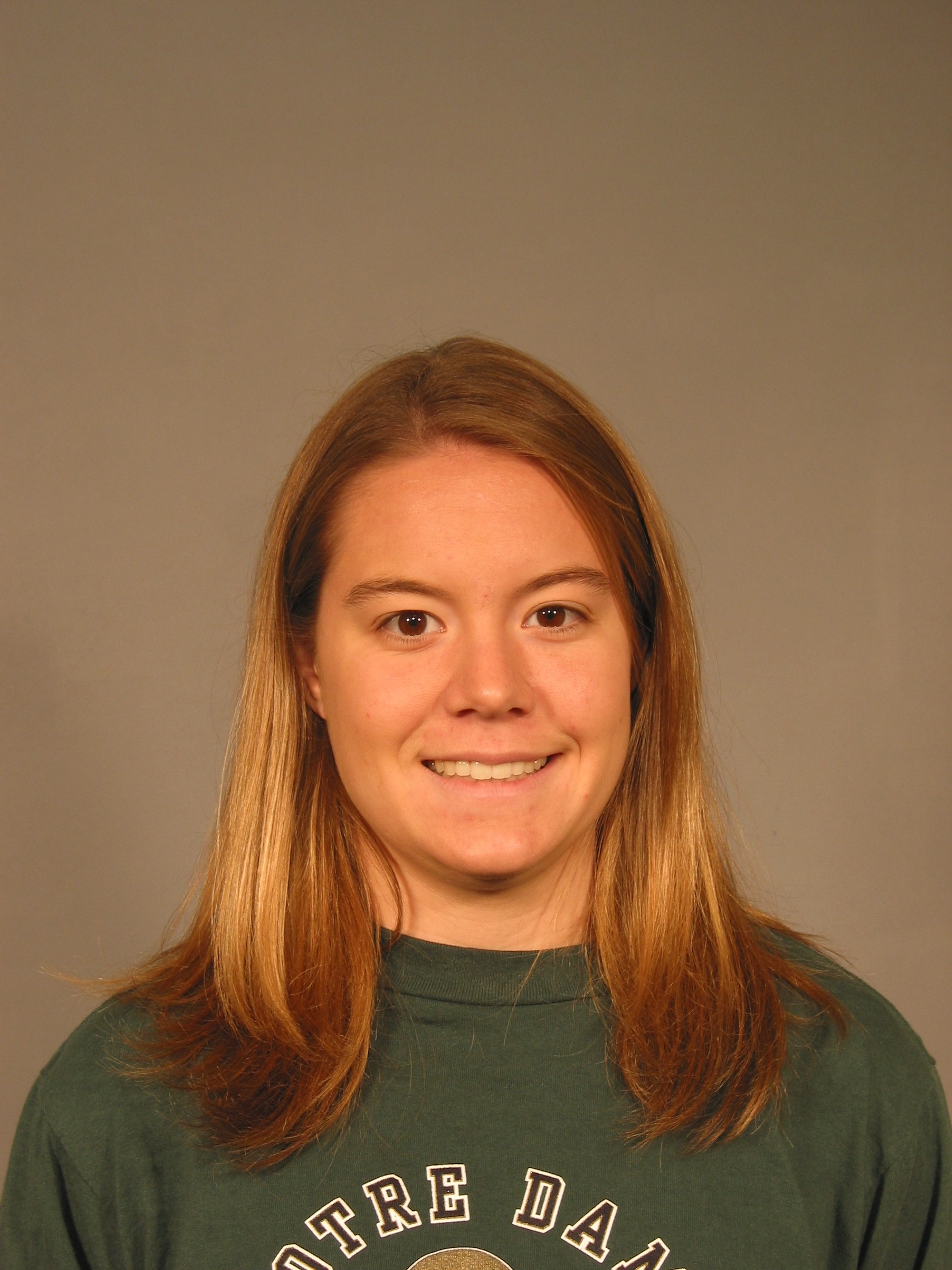}
}\end{parbox}
\end{tabular}
\end{small}
\end{center}
 \caption{
  Left plane: Non-cooperative traveler (04408d35) waiting for service in front of an ABC machine;  Right plane: Cooperative traveler (04408d33) following the controlled face acquisition at the machine  (subject from the FRGC database \cite{FRGC}).
}\label{fig:photosFRGC}
  \end{figure}

\begin{figure}[!hbt]
\begin{center}
\begin{small}
\begin{tabular}{c} 
\includegraphics[width=0.35\textwidth]{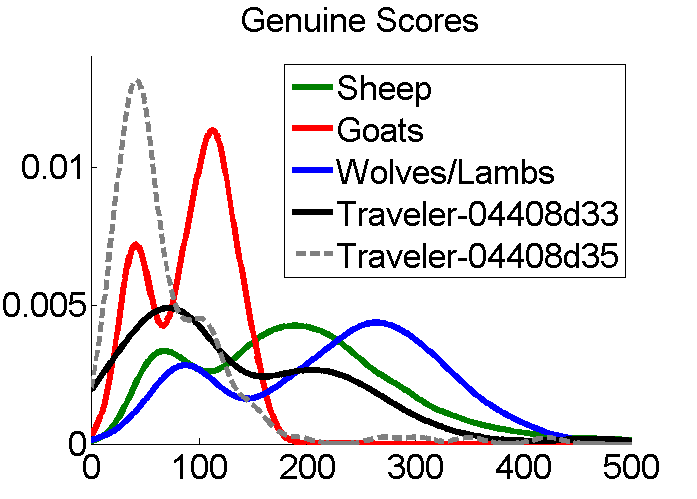}
\end{tabular}
\end{small}
\end{center}
 \caption{The probability density function for traveler 04408 from the FRGC database over Doddington's landscape.}\label{fig:KL-04408}
  \end{figure}

For example,  consider the evidence of non-cooperative and cooperative traveler in Fig. \ref{fig:photosFRGC}.
The  risks of traveler 04408 (which is on the watchlist) are evaluated in the KL metric over Doddington's landscape. In addition to the Doddington's categories, we plot the KL measures of the traveler 04408 while being non-cooperative (04408d35), and cooperative (04408d33), in Fig. \ref{fig:KL-04408}. We observe that:

1)  The image of a non-cooperative traveler (04408d35) will yield a probability density function with a spike located near the match score of 50.
	
2)  The image of a cooperative traveler (04408d33) will yield a probability density function that is more spread between all the possible match scores with a peak at a match score of 90.

In this paper, we extend the applications of relative entropy to the analysis of risk landscape for a watchlist check. 
The KL divergence is used to measure the difference between two probability density functions.  In this scenario, the KL measure of difference can be used to approximate the probability of a traveler belonging to the different types of Doddington's categories.
In terms of Bayes risk in a  decision making system, ``an expected loss is called a risk'' \cite{Duda}:  
	\[
	R(kl|x)=\sum^{t}_{j=1}\lambda(kl|w_j)D(w_j|x)
\]
where $t$ is the number of Doddington's categories, $R(kl|x)$ is the conditional risk of using KL decision ($kl$) given the probability density function of a person ($x$), $\lambda(kl|w_j)$ is the loss of using KL decision ($kl$) for the Doddington's categories ($w_j$)  and $D(w_j|x)$ is the KL divergence between $w_j$ and $x$.

Table \ref{tab:KL2} contains the KL measure calculated for particular traveler in each category with respect to the average (prior) distribution of score for the classes. For example, the risk of traveler 4408 is calculated as follows:
\begin{multline}\label{eq:Traveler-4408}
	R(kl|x_{4408})=\lambda(kl|w_s)D(w_s|x_{4408})\\ + \lambda(kl|w_g) D(w_g|x_{4408}) +\lambda(kl|w_wl)D(w_wl|x_{4408})
\end{multline}

Table \ref{tab:KL2} contains the KL measure calculated for particular traveler in each category with respect to the average (prior) distribution of score for the classes.

\begin{table}[!htb]
	\centering
		\caption{Relative entropy ($D(w_j|x)$) between 'sheep' and other Doddington categories using the FRGC database.}
		\begin{footnotesize}
		\begin{tabular}{|c|c||r|r|r|r|}
		\hline
		 & \multicolumn{1}{|c||}{Traveler} & \multicolumn{1}{|c|}{'Goats'} & \multicolumn{1}{|c|}{'W/L'} & \multicolumn{1}{|c|}{'Sheep'} & \multicolumn{1}{|c|}{$R(kl|x)$}\\
		\hline
		\hline
		\parbox[t]{2mm}{\multirow{8}{*}{\rotatebox[origin=c]{90}{'Goats'}}}  & \multicolumn{5}{|c|}{ Genuine Scores} \\
\cline{2-6}
&	4315	&	19.3834	&	1.5000	&	1.2857	&	12.2086	\\
&	4493	&	21.4335	&	1.4057	&	1.2061	&	13.4024	\\
&	4472	&	 8.1941	&	1.2029	&	1.0842	&	 5.3858	\\
\cline{2-6}
& \multicolumn{5}{|c|}{ Imposter Scores} \\
\cline{2-6}
&	4315	&	4.9675	&	1.9530	&	4.7111	&	4.0375	\\
&	4493	&	4.7866	&	1.8759	&	4.5387	&	3.8886	\\
&	4472	&	7.2058	&	2.8461	&	6.8617	&	5.8635	\\
\hline
\hline
\parbox[t]{2mm}{\multirow{8}{*}{\rotatebox[origin=c]{90}{'Wolves/Lambs'}}} & \multicolumn{5}{|c|}{ Genuine Scores} \\
\cline{2-6}
&	4202	&	75.8104	&	1.3291	&	0.3451	&	45.9195	\\
&	4408	&	44.6677	&	1.1112	&	0.3347	&	27.1675	\\
&	4435	&	84.4652	&	1.2451	&	0.0621	&	51.0589	\\
\cline{2-6}
& \multicolumn{5}{|c|}{ Imposter Scores} \\
\cline{2-6}
&	4202	&	4.7009	&	1.8339	&	4.4565	&	3.8164	\\
&	4408	&	4.9644	&	1.9450	&	4.7082	&	4.0330	\\
&	4435	&	4.6357	&	1.8120	&	4.3935	&	3.7644	\\
\hline
\hline
		\parbox[t]{2mm}{\multirow{8}{*}{\rotatebox[origin=c]{90}{'Sheep'}}} & \multicolumn{5}{|c|}{ Genuine Scores} \\
		\cline{2-6}
&	2463	&	38.2000	&	1.3801	&	0.8165	&	23.4157	\\
&	4201	&	55.8247	&	1.1022	&	0.2087	&	33.8464	\\
&	4203	&	42.4947	&	0.8047	&	0.3470	&	25.7729	\\
\cline{2-6}
& \multicolumn{5}{|c|}{ Imposter Scores} \\
\cline{2-6}
&	2463	&	8.4654	&	3.3284	&	8.0840	&	6.8862	\\
&	4201	&	7.0656	&	2.7951	&	6.7258	&	5.7505	\\
&	4203	&	7.3538	&	2.9030	&	7.0048	&	5.9837	\\
		\hline
		\end{tabular}
		\end{footnotesize}
		\label{tab:KL2}
\end{table}

Based on a typical security scenario, we can assume the following:

1) 'Sheep' have low risk because they do not cause many mis-identifications or rejections,
	
2) 'Wolves/Lambs' have medium risk because they only impact false accepts, which may cause false alarms regarding regular travelers being mistaken for the wanted persons, while the persons on the watchlist are unlikely to be missed, as they would rather produce multiple matches against the watchlist, and
	
3) 'Goats' have high risk because they  cause false rejects, which may cause a passage of a wanted person whose current appearance at the border checkpoint was not matched against his/her  facial image on the watchlist.

In this given scenario, we'll adopt the following loss values:
\[
\left [\hspace{-1.5mm}
\begin{array}{ll}
\lambda(kl|w_s)&\hspace{-2.5mm}=\\
\lambda(kl|w_g)&\hspace{-2.5mm}=\\
\lambda(kl|w_w/l)&\hspace{-2.5mm}=\\
\end{array}%
\begin{array}{ll}
\hspace{-2.5mm}0.1\\
\hspace{-2.5mm}0.6\\
\hspace{-2.5mm}0.3\\
\end{array}%
\hspace{-1.5mm}\right]
\begin{array}{ll}
\texttt{Low \hspace{1.5mm}Risk}\\
\texttt{High \hspace{1.5mm}Risk}\\
\texttt{Medium \hspace{1.5mm}Risk}\\
\end{array}%
\hspace{-1.5mm}
\]

It follows from this taxonomical view that:

\begin{itemize}
	\item [] \hspace{-7mm}\textbf{Low watchlist check risk}   is assigned to the travelers who do not impact the Doddington landscape or their impact is negligible. 
	\item  []\hspace{-7mm}\textbf{Medium watchlist check  risk} is assigned to the travelers who are wrongly detained, and, therefore, are costing additional resources to resolve (officer's time and suspect's discomfort).  
	\item []\hspace{-7mm}\textbf{High watchlist check risk}  is assigned to the travelers who have been identified as ``wanted``,  but nevertheless, have gained access to facility.
\end{itemize}  

Finally, the total risk of watchlist screening for Traveler 4408 is calculated based on Equation \ref{eq:Traveler-4408}:
\begin{multline*}
	R(kl|x_{4408})=0.6\cdot 44.6677\\ + 0.3\cdot 1.1112 + 0.1\cdot 0.3347=27.1675
\end{multline*}

In the last column of Table \ref{tab:KL2}, the $R(kl|x)$ represents the calculated risk for each traveler using the relative entropy.  Since the calculated KL values for 'goats' is a much larger value than the other categories, 'goats' contribute a greater portion of the calculated risk.  In this scenario, the risk value can be interpreted as the similarity between a traveler and a 'goat' class.  A high $R(kl|x)$ indicates a close similarity to the 'wolves/lambs' category, a low $R(kl|x)$ represents a closeness to the 'goat' category and a medium $R(kl|x)$ expresses a close proximity to the 'sheep' category.  For example, a $R(kl|x)=5.3858$ (genuine score for traveler 4472) indicates that the traveler belongs to the 'goat' category and a $R(kl|x)=45.9195$ represents a 'wolf/lamb' category.   

Some methods of risk assessment use cost evaluation as a way to numerically represent the gains and loss in the event of a crisis.  In general, Equation \ref{eq:cost} is proposed to assess the expected cost. This paper extends this concept and suggests to apply Equation \ref{eq:cost} which not only includes basic FNR and FPR error rates but also incorporates these error rates to the Doddington categories allowing for more specific cost estimation.

\subsection{Comparison of cost and relative entropy metric}

The multiparametric cost and relative entropy metric introduces different views of risk landscape.  The cost metric uses various parameters for controlling the factors of risk. For example, security personnel can increase or decrease the alarm level depending on various social or scenario factors. The relative entropy metric shows the relationships of different risk patterns of the watchlist landscape. These control  indicators convey proactive information on watchlist use, for example, after updating. Let us consider, in  more detail, the comparison of these metrics: 

1) The  multiparametric cost metric offers the control parameters of $T$, $C_{FN}$ and $C_{FP}$, and  generates eight functions of cost ('sheep', 'goat', 'wolf/lamb' and 'worm' categories with two cost function per category).  Since $FNR$ only impacts 'goat' and 'worm', and $FPR$ only influences 'wolf/lamb' and 'worm', only four of the eight cost functions are being considered.  In summary, the cost metric provides four cost functions that can be used to calculate the risk associated for each Doddington category at a specific threshold.
	
2) The relative entropy metric compares the historical data of match scores associated with each Doddington category with a selected traveler's score information to generate a similarity metric.  By finding the product of the relative entropy metric and the loss value associated with the Doddington category, it allows for the estimation of a numerical risk value of a selected traveler based on how similar his/her match score is to a Doddington category.  In addition, these metrics can be further analyzed by machine learning methods such as convolutional neural networks or fuzzy logic to obtain better association between a traveler and their Doddington categories.

\section{Discussion and conclusions}

The risk detectors proposed in this paper use both the Doddington categories of impersonation phenomenon, and the information-theory approach called Kullback-Leibler measures.  We experimentally demonstrate the effects of mis-identification and impersonation under various watchlist scenarios and constraints, and measure their impact, in terms of cost and risks, on the border control system performance. 

Despite impressive results of pattern recognition in forensic applications and entry-exit systems used only for visa-travelers, the facial watchlists in identification mode are still not present in border crossing applications, due to both technological and logistics challenges \cite{[EC-SmartBorders-2014]}. Addressing these challenges, our study suggests as follows:
\begin{enumerate}
	\item  Doddington metrics shall be used to categorize a watchlist to account for the unstable nature of a watchlist system.
	\item Scenario-oriented  multiparametric cost function shall be derived for evaluation of the risks of impersonation. The proposed cost function  is adapted to the specific scenario of watchlist screening by the set of recognition related indicators. This means that we can control the overall unstable Doddington landscape by its performance parameters ($T$, $FRR$, $FAR$), as well as security related parameters, $C_{FN}$, and $C_{FP}$.  
			\item Relative entropy metric for evaluating the potential risk of the watchlist screening for a given traveler with respect to Doddington categories. It is reasonable to incorporate entropy measures into Bayes risk model because it offers the possibility for numerically estimation risks of watchlist check associated with each traveler. In addition, this risk value can be controlled by the loss parameter ($\lambda(kl|w)$) that is unique to each Doddington category.  
			
\end{enumerate}

\subsection*{Acknowledgment}

\begin{small}
This project was partially supported by  Natural Sciences and Engineering Research Council of Canada (NSERC) through grant ``Biometric intelligent interfaces''; and  the Government of the Province of Alberta (ASRIF grant and Queen Elizabeth II Scholarship).
\end{small}

    {\small
\bibliographystyle{IEEEtran}
\bibliography{egbib}
}

\end{document}